\documentclass[mathematics,article,preprints,pdftex,moreauthors]{Definitions/mdpi}
\renewcommand{\linenumbers}{}
\usepackage{graphicx}
\usepackage{algorithm}
\usepackage{algpseudocode}
\usepackage{subfig}

\firstpage{1} 
\pubvolume{1}
\issuenum{1}
\articlenumber{0}
\pubyear{2023}
\copyrightyear{2023}
\datereceived{ } 
\daterevised{ } 
\dateaccepted{ } 
\datepublished{ } 
\hreflink{https://doi.org/} 
\pdfoutput=1 

\Title{Improving the performance of object detection by preserving label distribution}

\TitleCitation{Improving performance of object detection by preserving label distribution}

\Author{Heewon Lee, Sangtae Ahn*}

\AuthorNames{Heewon Lee, Sangtae Ahn}

\AuthorCitation{Lee, H; Ahn, S}

\address[1]{School of Electronic and Electrical Engineering, Kyungpook National University, 80 Daehak-ro, Buk-gu, Daegu, South Korea, 41566}

\corres{Correspondence: Sangtae Ahn (stahn@knu.ac.kr)}

\abstract{
Object detection is a task that performs position identification and label classification of objects in images or videos. The information obtained through this process plays an essential role in various tasks in the field of computer vision. In object detection, the data utilized for training and validation typically originate from public datasets that are well-balanced in terms of the number of objects ascribed to each class in an image. However, in real-world scenarios, handling datasets with much greater class imbalance, i.e., very different numbers of objects for each class , is much more common, and this imbalance may reduce the performance of object detection when predicting unseen test images. In our study, thus, we propose a method that evenly distributes the classes in an image for training and validation, solving the class imbalance problem in object detection. Our proposed method aims to maintain a uniform class distribution through multi-label stratification. We tested our proposed method not only on public datasets that typically exhibit balanced class distribution but also on custom datasets that may have imbalanced class distribution. We found that our proposed method was more effective on datasets containing severe imbalance and less data. Our findings indicate that the proposed method can be effectively used on datasets with substantially imbalanced class distribution.
}


\begin{document}

\section{Introduction}

Computer vision is a field of artificial intelligence that enables machines to understand and interpret visual information \cite{cv-survey}. Computer vision involves the recognition and extraction of patterns from images or videos, and it has been applied in various fields \cite{cv_app1,cv_app2,cv_app3,cv_app4,cv_app5,cv_app6,cv_app7,cv_app8,cv_app9,cv_app10,cv_app11,cv_app12,cv_app13,cv_app14,cv_app15,cv_app16}. Examples of its usage include object detection \cite{od_survey}, image classification \cite{class_survey}, face recognition \cite{face_survey}, or image generation \cite{img_gan_survey}. Among these, object detection is an important task in the field of computer vision, which aims to identify and localize multiple objects in images or videos.

To develop a model for object detection, the first step is to collect data and then appropriately split them into training and validation datasets. When splitting the original data, maintaining a similar class distribution between the training and validation datasets is crucial because datasets for object detection are presented as multi-label problems in that multiple objects can be present in a single image. In particular, the number of objects for each label could be imbalanced (Figure 1). Maintaining a similar class balance when dividing the dataset could be difficult. For image classification problems, a strategy called stratification can be used to maintain similar class distributions. However, it is unknown how stratification is applied to an object detection problem and how stratification influences the perofmrance of object detection.

\begin{figure}
    \includegraphics[width=\textwidth]{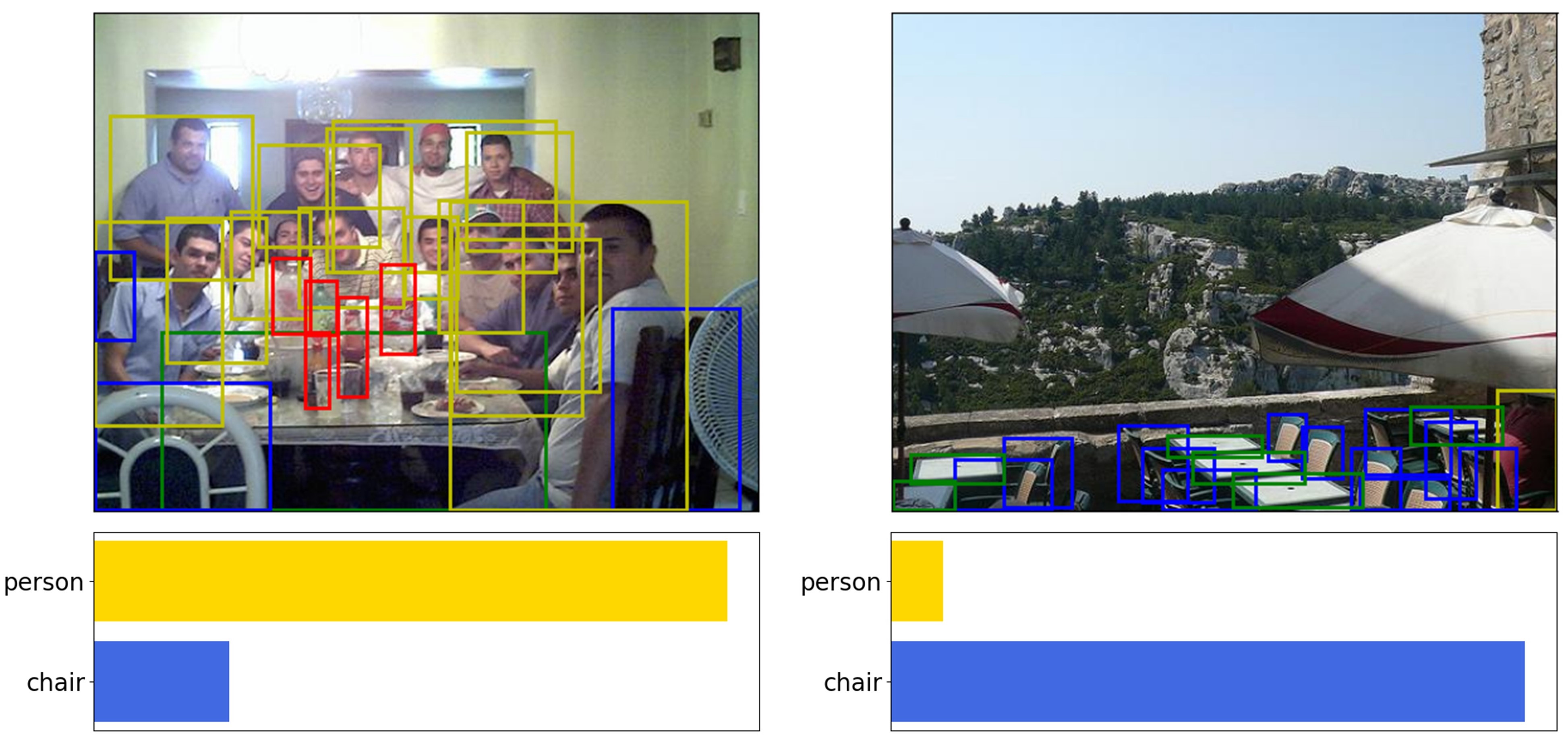}
    \caption{Each image has a different number of bounding boxes for each class. For example, the number of bounding boxes for a person may be high while the number of bounding boxes for a chair may be low (left) or vice versa (right). Images are taken from the Pascal Visual Object Classes(VOC) 2012 dataset.}
    \label{fig}
\end{figure}

Here, we present a study that proposes a new method called stratification for object detection (SOD), which applies stratification to the object detection problem. This method uses a multi-label stratification technique to preprocess labeled data in object detection and then applies stratification. This method facilitates the preservation of the class distribution among the split datasets, yielding better performance on public and custom datasets. Our proposed method can solve the problem of class imbalance better than existing methods, improving the performance of trained models. Our method was applied to object detection problems, particularly implemented in the YOLO format because the YOLO algorithm is a widely used platform in object detection studies. Therefore, our proposed method is expected to be a useful tool that solves an important problem in the field of object detection and guarantees improved performance. Furthermore, it can contribute to improving the accuracy and reliability of the object detection model.

The main contributions of this research are as follows:
\begin{itemize}
    \item We propose a method for preserving class distribution in object detection tasks.
    \item We experimentally demonstrate the effectiveness of our proposed method on both public datasets and custom datasets.
\end{itemize}

\section{Related Work}
\subsection{Real-time object detection}
Object detection is a computer vision task that involves detecting multiple objects in images or videos and classifying their positions and types. It is a key technological tool that enables computers to understand the real world and has been applied in various fields such as autonomous driving\cite{od_app1}, surveillance systems\cite{od_app2}, robotics\cite{od_app3}, augmented reality\cite{od_app4}, and medical image analysis\cite{od_app5}. Fundamentally, object recognition involves representing specific objects in an image as rectangular bounding boxes and characterizing them into different classes. While there are various algorithms for this task, we mainly employ the YOLO algorithm in this study.

YOLO\cite{yolov1}, first proposed in 2015, is an algorithm that enables real-time object recognition. YOLO divides an image into a grid and applies a method to simultaneously predict bounding boxes and class probabilities for each grid cell. Unlike two-stage detectors\cite{to_stage}, YOLO adopts a one-stage detector approach, whereby it considers the entire image at once and performs predictions. This unique feature of YOLO allows for real-time processing. Since 2015, YOLO has undergone many improvements. In YOLOv2\cite{yolov2}, the first update of YOLO, the ability to detect objects of various sizes was enhanced. The concept of anchor boxes\cite{anchor_box} was introduced in YOLOv2, and multi-scale training methods were used to train on images with different resolutions, resulting in improved detection of objects of different sizes. Additionally, YOLOv2 utilized the WordTree model to jointly train on the MSCOCO and ImageNet datasets, enabling the detection of over 9,000 different classes. YOLOv3\cite{yolov3} improved the detection of objects of various sizes by predicting bounding boxes in three different-sized feature maps. YOLOv3 also introduced a method to predict multi-labels for each box, facilitating the handling of more complex classification problems. In YOLOv4\cite{yolov4}, several optimizations were introduced to improve both performance and speed over those of previous versions. For YOLOv4, several features were introduced, namely, a new backbone network called CSPDarknet53, a new neck structure using PANet and SAM blocks, and the Mish activation function. These new structures and features made YOLOv4 more efficient and capable of accurately recognizing objects. The improved training speed, inference time, and model size were improved in YOLOv5\cite{yolov5}, while user-friendly features were added using the PyTorch framework. This enabled faster and more efficient object recognition, expanding the practical application of object detection. Recently, the trainable bag-of-freebies method was introduced in YOLOv7\cite{yolov7,bof} to significantly improve detection accuracy without increasing the inference cost. In this version, methods were also proposed to address issues arising from re-parameterized modules replacing original modules and apply dynamic label assignment strategies for different output layer assignments. Furthermore, extended and compound scaling methods were formulated to effectively utilize parameters and computations, reduce parameters and computational cost, and improve inference speed and accuracy.
In this manner, YOLO-based models have continued to evolve and become the standard in real-time object detection. By examining these research trends, we can observe that object detection algorithms are steadily advancing to become faster, more accurate, and more applicable in diverse environments.

\subsection{Stratification}
Studies applying stratification to datasets have emphasized the significance of the class distribution of the dataset, which is particularly important in the treatment of classification problems in the field of machine learning\cite{stratified_classfier}. Stratification is a type of data sampling technique that seeks to maintain the class distribution of the entire dataset in the training and validation sub-datasets. The key advantage of this technique is that it ensures that all classes are represented in the training and validation subsets as they are over the entire dataset, allowing the model to earn each class fairly.

The utilization of stratification in deep learning is mainly focused on addressing class imbalance issues\cite{class_imbalance_stratified}. In scenarios in which the number of samples for a specific class is significantly larger than that for other classes, i.e., class imbalance, the performance of the model can be excessively influenced by the dominant class. Stratification can alleviate such problems and ensure that all classes are fairly represented.

Furthermore, applying stratification to cross-validation is a highly effective strategy. Cross-validation\cite{cross_validation} is a technique used to evaluate the generalization performance of a model by dividing the data into multiple folds and using each fold alternately as a training set and a validation set. By applying stratification to this process, the class distribution of the data in each fold can accurately reflect the distribution of the entire dataset. This can help verify whether each model performs equally well for all classes.

Various methods for applying stratification in deep learning have been utilized, and they can be adjusted depending on the characteristics of a specific problem and dataset. Stratification may be necessary in some cases but non-essential in others. However, in general, stratification is recognized as an important step for improving model training and enhancing generalization performance.

\subsection{Multi-label stratification}
Multi-label classification\cite{multi_label_survey}, unlike conventional classification, allows each data point to be classified with multiple labels simultaneously. Stratified sampling is a technique that selects samples evenly from each category, ensuring the representativeness of the entire dataset while maintaining the importance of each category. However, this technique is only applicable to single-label classification problems. Multi-label stratification extends this technique to problems in which each data point can have multiple labels. This ensures that each label combination is evenly distributed in the training and validation datasets, allowing the model to optimize generalizability for each label combination during training.

There are various use cases for multi-label stratification. For example, in music classification\cite{multi_label_music}, as a song can belong to multiple genres, the corresponding training and validation datasets should represent each genre combination correctly to achieve accurate model learning. Multi-label stratification can be used to meet these requirements. Similarly, in medical imaging\cite{multi_label_medical}, an image can display multiple diseases. In this case, the combinations involving each disease label need to be well represented in the training and validation datasets, and multi-label stratification can be used for this purpose. Additionally, in text classification\cite{multi_label_text}, text data such as news articles, research papers, and blog posts can be simultaneously classified multiple topics or categories. In this case, multi-label stratification can be applied to ensure that combinations involving each topic are evenly distributed in the training and validation datasets. In summary, multi-label stratification plays an important role in enabling more accurate model training in various fields to address real-world problems effectively.

\section{Proposed algorithm}

\begin{algorithm}[!t]
\caption{YOLOstratifiedKFold}
\begin{algorithmic}[1]
\renewcommand{\algorithmicrequire}{\textbf{Input:}}
\Require $F_{img},\ F_{txt},\ k$

\State $L_{img} \gets$ empty list \Comment{List of image file names}
\For{$f$ in $listdir(F_{img})$ with $.jpg$ extension}
    \State append $splitext(f)_0$ to $L_{img}$
\EndFor
\State $L_{img} \gets$ remove duplicates from $L_{img}$
\vspace{11pt}

\State $L_{txt} \gets$ empty list \Comment{List of txt file names}
\For {$f$ in $listdir(F_{txt})$ with $.txt$ extension}
    \State append $splitext(f)_0$ to $L_{txt}$
\EndFor

\State $L_{txt} \gets$ remove duplicates from $L_{txt}$ 
\vspace{11pt}

\State $L_{data} \gets$ empty list \Comment{List of files data}
\For {each $F_{name}$ in $L_{img}$} 
    \If{$F_{name}$ is not in $L_{txt}$}
        \State $L_{data} \gets$ append $[F_{name}+'.jpg',-1,None,None,None,None]$
    \Else
        \State $txt\_file\_path \gets$ join $F_{txt},\ F_{name}\ +\ '.txt'$
        \State $f \gets$ open $txt\_file\_path$
        \State $lines \gets$ read all lines from $f$
        \If{$lines$ is not empty}
            \State $L_{data} \gets$ append $[F_{name}\ +\ '.jpg',-1,None,None,None,None]$
        \Else
            \For {each $line$ in $lines$}
                \State $L_{data} \gets$ append $[F_{name}\ +\ '.jpg',line_0,line_1,line_2,line_3,line_4]$
            \EndFor
        \EndIf
    \EndIf
\EndFor
\vspace{11pt}

\State $data \gets$ create Dataframe from $L_{data}$
\vspace{11pt}

\State $one\_hot \gets$ convert $data['class']$ to one-hot encoding
\State $data \gets$ concatenate $data$ and $one\_hot$ and multiply 1000 to 'w', 'h' columns
\State $new\_df \gets$ drop 'class', 'x', 'y' columns from $data$ and group by 'filename' and sum
\State $new\_df_{cnt} \gets$ replace 0 with 1 in count of box
\State $new\_df_{avg\_w}, new\_df_{avg\_h}, new\_df_{avg\_ratio} \gets$ Calculate average width,  height, ratio
\State drop 'w','h' columns from $new\_df$
\vspace{11pt}

\State $mskf \gets$ initialize $\textbf{MultilabelStratifiedKFold}$ with $k$
\vspace{11pt}

\For {each $(train\_idx, val\_idx)$ in $mskf.split(new\_df_{filename},\ new\_df.iloc[:,1:])$}
\State $X\_train, X\_val \gets$ select rows from $new\_df$ by $train\_idx, val\_idx$
\EndFor

\end{algorithmic}
\end{algorithm}
In this section, we introduce a new algorithm that applies stratification to object detection datasets using the YOLO format. The main goal is to improve existing dataset partitioning methods, enabling a more precise and fairer training process. The detailed operation of the algorithm is as follows.

This algorithm requires three inputs: the paths to the folders where the image files and text files are stored, denoted as $F_{img}$ and $F_{txt}$ respectively, and the number of subsets within the dataset to be created, denoted as $k$.

The algorithm, illustrated below, operates as follows. Lines 1 to 5 generate a list using the names of the image files. Similarly, lines 6 to 10 construct a list using the names of the text files. Lines 11 to 27 convert the label data from the text files into Dataframe format. In this process, if the dataset contains background images to prevent false positives, there may be no corresponding text file for that image or the text file may be empty to prevent the presence of false positives. To account for such background images in the partitioning process, we insert -1 in the class column of the Dataframe. Once this step is completed, the Dataframe contains the paths of the text files, number of objects per class, and the x and y coordinates, width (w), and height (h) of each object. Lines 29 to 34 are preprocessing steps needed before the multi-label stratified KFold procedure is performed. In this process, we perform one-hot encoding on the classes and concatenate the original Dataframe with the one-hot encoded Dataframe. We then multiply the $w$ and $h$ values by 1000 to prevent loss of values when calculating their averages. Next, we remove the "class", "x", and "y" columns from the Dataframe and group them based on filename. This creates a Dataframe that indicates how many objects of each class exist within each text file. For background images, the total object count is 0; thus, we change it to 1 to avoid division by 0 when calculating the averages of $w$ and $h$. We then calculate w, h, and the ratio of h to w. Then, we remove the "w" and "h" columns to improve computational efficiency during the partitioning process. Finally, lines 35 to 38 perform the multi-label stratified KFold procedure, dividing the dataset into training and validation sets.

We discuss the selection of the labels for the Dataframe in the multi-label stratified KFold process $class_0\sim class_n$, $avg\_w$, $avg\_h$, and $avg\_ratio$. $class_0\sim class_n$ indicates the presence and quantity of each class within an image. This parameter serves to ensure diversity in the training data by considering the various class labels within the dataset. We also selected average width ($avg\_w$), average height ($avg\_h$), and average ratio ($avg\_ratio$) as labels for the following reason. Object detection models such as Faster R-CNN\cite{fastrcnn} and YOLO use predefined anchor boxes with specific aspect ratios. Anchor boxes are one of the elements used in object detection, in which frames with specific positions and sizes are set within an image. This allows the model to detect objects of various sizes and shapes, resulting in improved accuracy. If the aspect ratio distribution of bounding boxes in the training set and validation set differs, model performance can be negatively affected. If the aspect ratio distribution is not aligned, it becomes difficult to appropriately set the position and size of the anchor boxes, which ultimately affects the accuracy of object detection. Therefore, by selecting the average width ($avg\_w$), average height ($avg\_h$), and average ratio ($avg\_ratio$) as labels, we aim to characterize the aspect ratio distribution of the bounding boxes in each dataset and optimize their size and aspect ratios, improving model performance. This allows the model to effectively detect objects with various aspect ratios.

\section{Experiments}
\subsection{Datasets}

We present the results of experiments conducted on public datasets to demonstrate the efficiency of the proposed algorithm. The datasets are used for object detection purposes, with each image indicating the location and type of object through a bounding box. Across the datasets, the number of samples is generally far greater than the number of classes, which enhances the training process by providing a wide range of samples for each class. In addition, we selected a dataset that includes background images to minimize the phenomenon of background false positives. This is crucial in preventing errors in object detection owing to features in the background.

We used entropy (Equation 1) to assess the distribution of each dataset. Higher entropy corresponds to higher uncertainty in the data and vice versa. Thus, entropy can be used as a measure of how well the classes in the data are distributed. In the equation, $p_i$ represents the occurrence probability of each class over the entire dataset.
\begin{equation}
    Entropy = - \sum_{i=1}^{n} p_{i} \log(p_{i})
\end{equation}

Table 1 shows a detailed analysis of the datasets. In datasets with a large number of classes, entropy tends to be relatively high. If the class label distribution of a dataset is diverse or uncertain, the stratification process may not be as effective. As stratification involves extracting samples while maintaining the class ratio of the original dataset, the proposed algorithm is recommended for use in scenarios in which the number of classes is less than or equal to 20.

\begin{table}[H]
\caption{Statistical analysis of datasets for object detection. Public datasets: COCOval2017, Pascal VOC 2012 val, and PlantDoc. Private datasets\cite{roboflow}: Website screenshot, Aquarium, and BCCD.}
    \begin{adjustwidth}{-\extralength}{0cm}
        \newcolumntype{C}{>{\centering\arraybackslash}X}
        \begin{tabularx}{\fulllength}{llcccCCCCCCc}
            \toprule
            \multirow{2}{*}{\textbf{Category}}	& \multirow{2}{*}{\textbf{Dataset}}	& \multirow{2}{*}{\textbf{Classes}}	& \multirow{2}{*}{\textbf{Samples}} & \textbf{Samples} & \multicolumn{3}{c}{\textbf{Samples per Image}}  & \multicolumn{3}{c}{\textbf{Class per Image}}  & \textbf{Entropy}\\
            \cline{6-8} \cline{9-11}
            & & & & \textbf{per Class} & Min & Avg & Max & Min & Avg & Max & \\
            \midrule
            \multirow{3}{*}{Public} & COCO val2017\cite{mscoco} & 80 & 5000 & 62.5 & 0 & 7.9 & 96 & 0 & 2.9 & 14 & 3.39 \\
            & Pascal VOC 2012 val\cite{pascalvoc2012} & 20 & 3422 & 171.1 & 0 & 2.3 & 23 & 0 & 1.4 & 5 & 2.31 \\
            & PlantDoc\cite{plantdoc} & 30 & 2569 & 85.6 & 0 & 3.4 & 42 & 0 & 1 & 3 & 3.17 \\
            \midrule
            \multirow{3}{*}{Private} & Website screenshot & 8 & 1206 & 150.8 & 2 & 45 & 2023 & 2 & 5.3 & 8 & 1.61 \\
            & Aquarium & 7 & 638 & 91.1 & 0 & 7.6 & 56 & 0 & 1.4 & 3 & 1.42 \\
            & BCCD & 3 & 364 & 121.3 & 1 & 13.4 & 30 & 1 & 2.5 & 3 & 0.53 \\
            \bottomrule
        \end{tabularx}
    \end{adjustwidth}
\end{table}

\subsection{Class distribution}

We applied 10-fold cross-validation to the datasets and compared the class distribution of the original dataset with that of the split datasets. To quantify the differences in distribution between them, we used the mean absolute error(MAE) (Equation 2). This calculation was used to measure the difference in class ratios between the original dataset and split dataset. Through this, we quantitatively evaluated how accurately the split dataset reflects the class distribution of the original dataset. In the MAE metric below, $y_i$ represents the class ratio of the original dataset, $\hat{y}_i$ represents the class ratio of the divided dataset, and $n$ represents the number of classes in the existing dataset.
\begin{equation}
    MAE = \frac{1}{n}\sum_{i=1}^{n}|y_i-\hat{y}_i|
\end{equation}

We compared the MAE of the traditional KFold cross-validation method and that of our proposed algorithm. As shown in Table 2, in the majority of datasets in which our proposed algorithm was applied, the median MAE was lower compared to that of the traditional KFold cross-validation. This demonstrates that our proposed algorithm was more effective in preserving the class ratios than the traditional method. However, some of datasets with an entropy of 2 or higher have found that KFold preserves class distribution better than our proposed algorithm. This indicates that such a phenomenon occurs when there is high uncertainty in the label distribution. In contrast, for datasets with an entropy of 2 or lower, our algorithm consistently showed lower median MAE and variance values compared to KFold in all cases. These results demonstrate that our algorithm provides a more stable and consistent performance.

Based on the experimental results, our proposed algorithm has been confirmed to effectively preserve class ratios while reducing variance, surpassing the commonly used KFold method.

\begin{table}[H]
\caption{Statistical characteristics of the subsets generated through 10-fold cross-validation. A comparison between KFold and our proposed algorithm is presented, listing the name of each dataset, entropy of each dataset, class distribution of the training set (9 folds), and class distribution of the validation set (1 fold). The unit of the MAE is 1e-7.}
    \begin{adjustwidth}{-\extralength}{0cm}
        \newcolumntype{C}{>{\centering\arraybackslash}X}
        \begin{tabularx}{\fulllength}{l|l|c|CC|CC}
            \hline
            \multirow{2}{*}{\textbf{Category}} & \multirow{2}{*}{\textbf{Dataset}} & \multirow{2}{*}{\textbf{Entropy}} & \multicolumn{2}{c|}{\textbf{Train}} & \multicolumn{2}{c}{\textbf{Validation}} \\
            \cline{4-5} \cline{6-7}
            & & & \textbf{KFold} & \textbf{Ours} & \textbf{KFold} & \textbf{Ours} \\
            \hline
            & COCO val2017 & 3.39 & \textbf{165$\pm $127} & 168$\pm $128 & \textbf{1466.5$\pm $1126.5} & 1506.5$\pm $1144.5 \\
            Public & Pascal VOC 2012 val & 2.31 & \textbf{591.5$\pm $408.5} & 618$\pm $391 & 5299$\pm $3712 & \textbf{5279$\pm $3330} \\
            & PlantDoc & 3.17 & 463.5$\pm $321.5 & \textbf{301.5$\pm $255.5} & 4097$\pm $2803 & \textbf{2614.5$\pm $2205.5} \\
            \hline
            & Website screenshot & 1.61 & 4864$\pm $3880 & \textbf{4092.5$\pm $3428.5} & 42897$\pm $34009 & \textbf{35538.5$\pm $29582.5} \\
            Private & Aquarium & 1.42 & 5172.5$\pm $4075.5 & \textbf{3943$\pm $2202} & 44031.5$\pm $33452.5 & \textbf{38541$\pm $22795} \\
            & BCCD & 0.53 & 1973.5$\pm $1417.5 & \textbf{1224$\pm $862} & 17683.5$\pm $12738.5 & \textbf{11459$\pm $8186} \\
            \hline
        \end{tabularx}
    \end{adjustwidth}
\end{table}

\subsection{Training and testing}

To analyze the difference in performance between KFold and our proposed algorithm, we applied 10-fold cross-validation using the two methods in question to split the dataset. The dataset used in our study was divided according to the following procedure (Figure 2).

\begin{enumerate}
\item The original dataset was split into 10 folds using our proposed algorithm.
\item The last fold (10th fold) was set as the test dataset.
\item The remaining 9 folds (k-1 folds) were combined and further split into 9 folds using KFold.
\item Similarly, the k-1 folds were split into 9 folds using our proposed algorithm.
\item Training was performed iteratively for each of the 9 fold datasets.
\item The trained models were used for inference on the fixed test dataset.
\item Finally, the inference results using KFold and our proposed algorithm were compared using MAE.
\end{enumerate}
The training was conducted with a 320-pixel image as the base model of YOLOv7 on a computing system consisting of three NVIDIA RTX 3090 Ti GPUs.

\begin{figure}[htb!]
    \includegraphics[width=\textwidth]{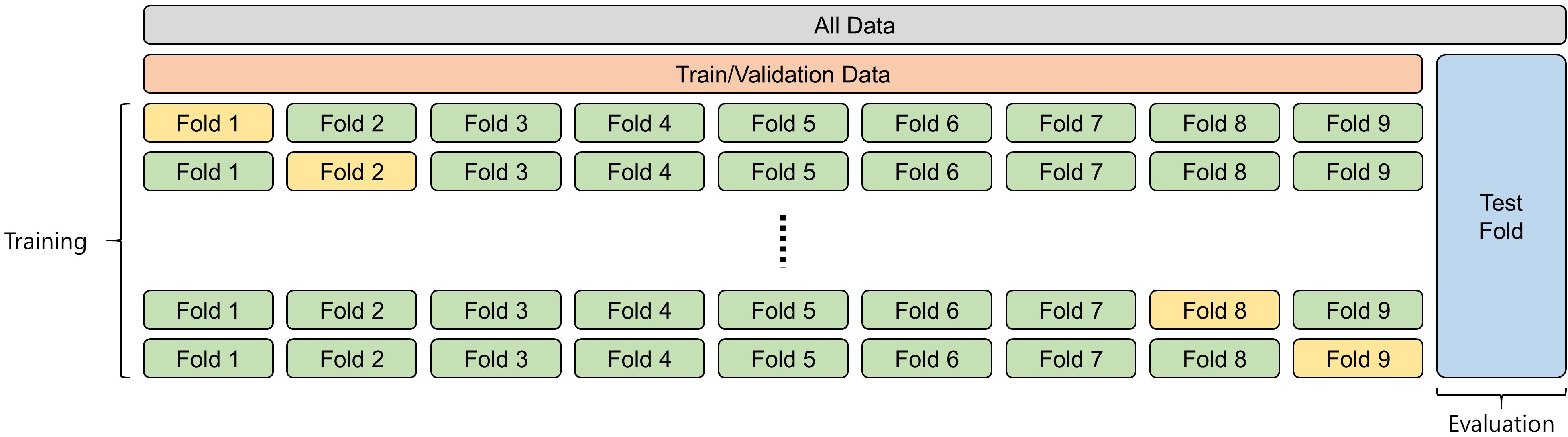}
    \caption{10-fold cross-validation split method. The yellow box represents the fold used for validation, and the green boxes represent the fold used for training. When splitting the Train/Validation data and Test data, YOLOstratifiedKFold is used, and when dividing the data onto folds, either KFold or YOLOstratifiedKFold is used.}
    \label{fig}
\end{figure}

For datasets in which the entropy value is 2 or less, the training on the dataset applied with our proposed method consistently showed higher performance in the $mAP\_0.5$ metric as compared to when the KFold method was used, illustrated in Table 3. In the custom datasets, which contain relatively few classes, the class distribution of the datasets applied with our proposed method was more similar to the original class distribution (refer to Table 2) than the public dataset, and we confirmed that the performance of our model was also higher than that of the original KFold method. These results suggest that in training with data that exhibit relatively low complexity, i.e., a low number of classes, our method can achieve higher performance based on the $mAP\_0.5$ metric.

\begin{table}[H]
\caption{Results of training by 10-fold cross validation}
    \begin{adjustwidth}{-\extralength}{0cm}
        \newcolumntype{C}{>{\centering\arraybackslash}X}
        \begin{tabularx}{\fulllength}{l|l|c|CC|CC}
            \hline
            \multirow{2}{*}{\textbf{Category}} & \multirow{2}{*}{\textbf{Dataset}} & \multirow{2}{*}{\textbf{Entropy}} & \multicolumn{2}{c|}{\textbf{mAP\_0.5}} & \multicolumn{2}{c}{\textbf{mAP\_0.5:0.95}} \\ 
            \cline{4-7}
            & & & \textbf{KFold} & \textbf{Ours} & \textbf{KFold} & \textbf{Ours} \\
            \cline{4-5} \cline{6-7}
            \hline
            \multirow{3}{*}{Public} & COCO val2017 & 3.39 & 23.50\%$\pm $0.80\%  & \textbf{23.75\%$\pm $1.25}\% & 14.00\%$\pm $0.80\% & \textbf{14.10\%$\pm $0.70\%} \\
            & Pascal VOC 2012 val & 2.31 & \textbf{46.05\%$\pm $2.25\%} & 44.95\%$\pm $2.85\% & \textbf{27.95\%$\pm $1.05\%} & 27.45\%$\pm $1.55\% \\
            & PlantDoc & 3.17 & \textbf{48.80\%$\pm $2.60\%} & 48.20\%$\pm $2.00\% & 32.90\%$\pm $1.90\% & \textbf{33.30\%$\pm $1.30\%} \\
            \hline
            \multirow{3}{*}{Private} & Website screenshot & 1.61 & 45.85\%$\pm $1.65\%  & \textbf{46.10\%$\pm $1.50\%} & \textbf{28.00\%$\pm $1.20\%} & 27.75\%$\pm $1.45\% \\
            & Aquarium & 1.42 & 51.60\%$\pm $3.20\% & \textbf{52.25\%$\pm $4.25\%} & 23.10\%$\pm $2.10\% & \textbf{23.25\%$\pm $2.25\%} \\
            & BCCD & 0.53 & 87.90\%$\pm $1.70\% & \textbf{88.55\%$\pm $1.35\%} & \textbf{55.55\%$\pm $1.25\%} & 55.45\%$\pm $1.55\% \\
            \hline
        \end{tabularx}
    \end{adjustwidth}
\end{table}

In contrast, it was difficult to confirm a significant improvement in the $mAP\_0.5:0.95$ metric from the training results. For our method, the datasets demonstrating better performance in $mAP\_0.5:0.95$ had  less than 100 samples per class. These results suggest that our method effectively works even with small datasets, demonstrating good performance even for low numbers of samples per class. This implies that when the dataset is split using our method, the model can recognize the rough location of the object more accurately; however, it does not reflect the detailed shape and exact location of the object during the splitting process. One possible way to solve this problem is to perform stratification by referring to the features within the bounding box in the image; however, this method significantly increases the computational load required for stratification, which could greatly increase the amount of time to perform the routine. Therefore, in this study, we minimized the time burden and verified the performance of the model by excluding image references and only using label data stored in text files.

\subsection{Inference results}

\begin{figure*}[p]
     \subfloat[(COCO val2017) KFold]{\includegraphics[width=0.48\textwidth]{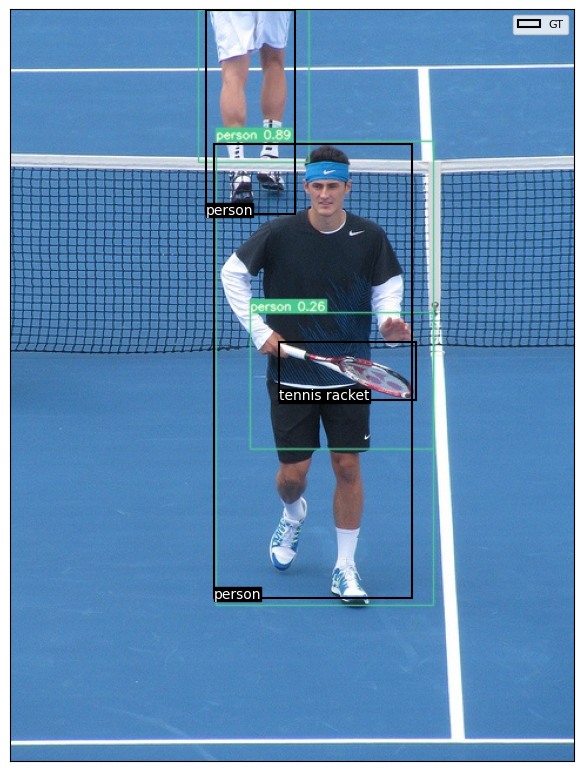}}
     \hfill
     \subfloat[(COCO val2017) Our method]{\includegraphics[width=0.48\textwidth]{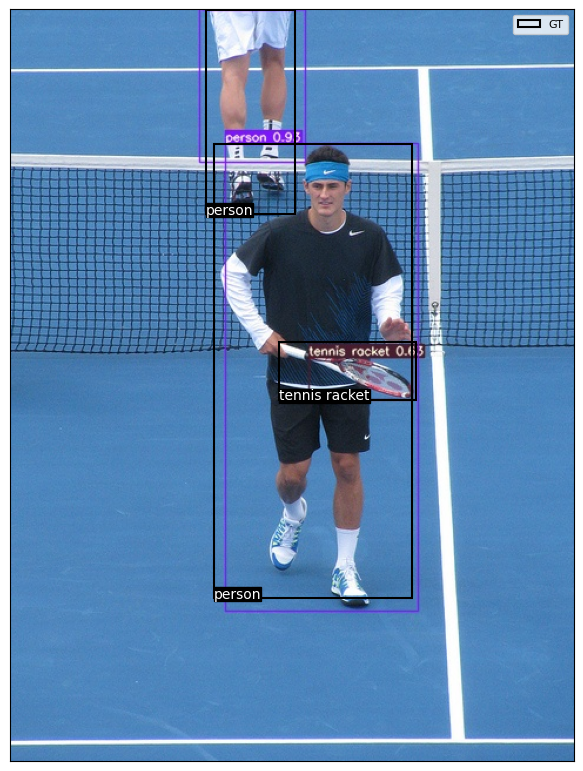}}
     
     \subfloat[(Pascal VOC 2012) KFold]{\includegraphics[width=0.48\textwidth]{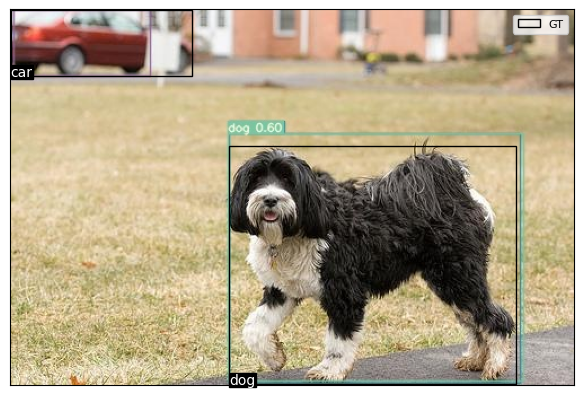}}
     \hfill
     \subfloat[(Pascal VOC 2012) Our method]{\includegraphics[width=0.48\textwidth]{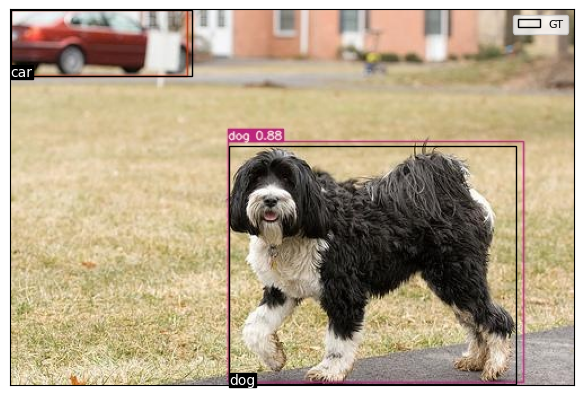}}
     
     \subfloat[(Plant) KFold]{\includegraphics[width=0.48\textwidth]{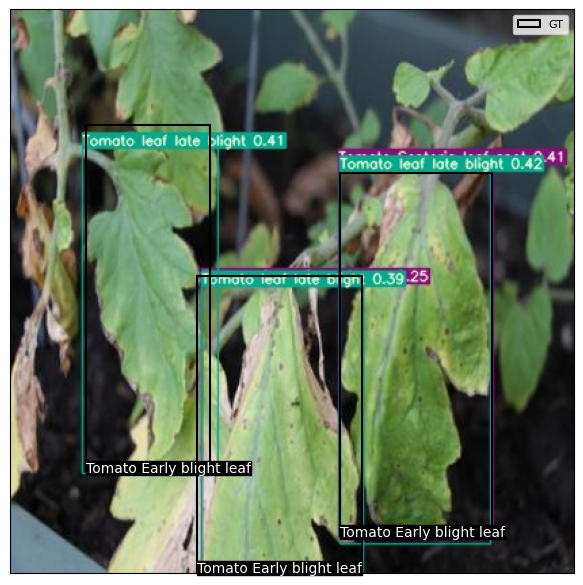}}
     \hfill
     \subfloat[(Plant) Our method]{\includegraphics[width=0.48\textwidth]{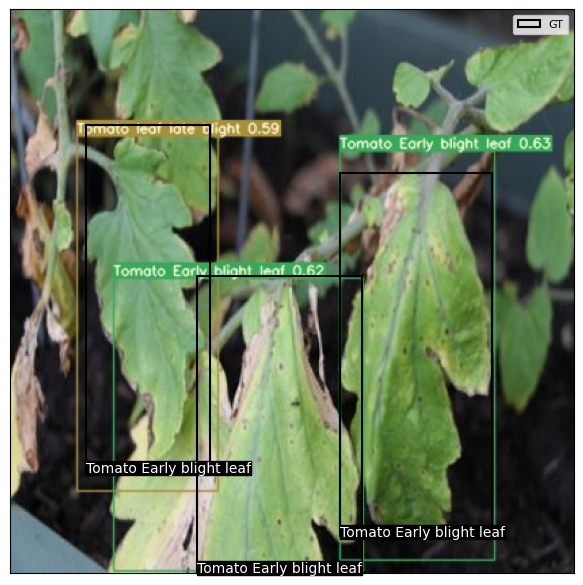}}
     \caption{Comparing KFold to our proposed method on the Public dataset.}
\end{figure*}

\begin{figure*}[p]
     \subfloat[(Website Screenshot) KFold]{\includegraphics[width=0.48\textwidth]{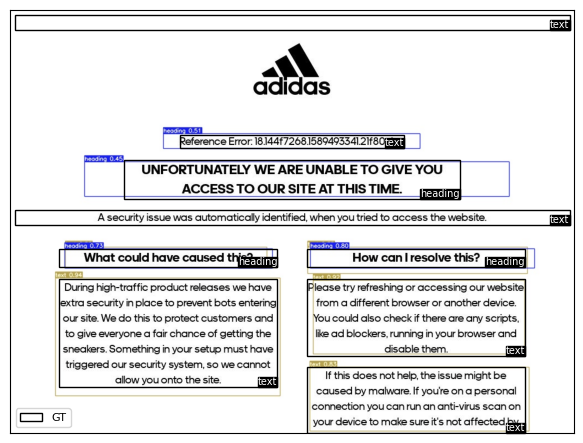}}
     \hfill
     \subfloat[(Website Screenshot) Our method]{\includegraphics[width=0.48\textwidth]{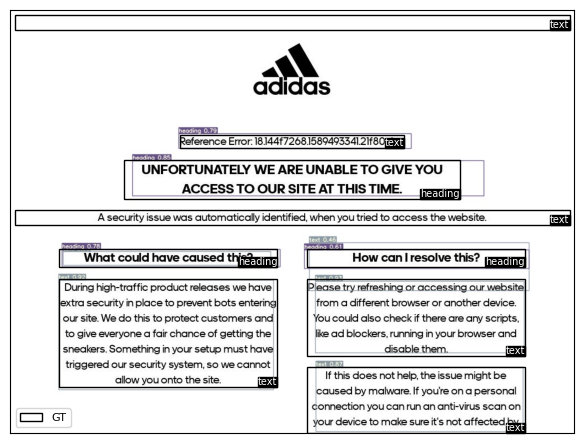}}
     
     \subfloat[(Aquarium) KFold]{\includegraphics[width=0.48\textwidth]{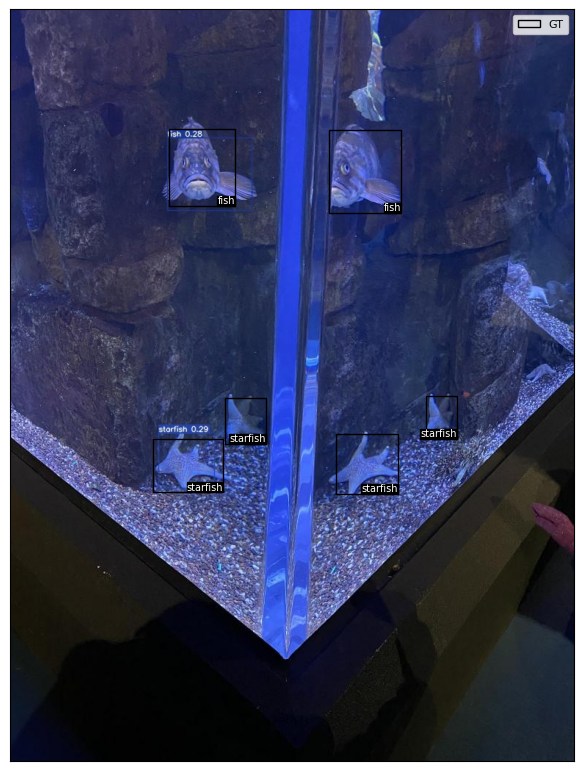}}
     \hfill
     \subfloat[(Aquarium) Our method]{\includegraphics[width=0.48\textwidth]{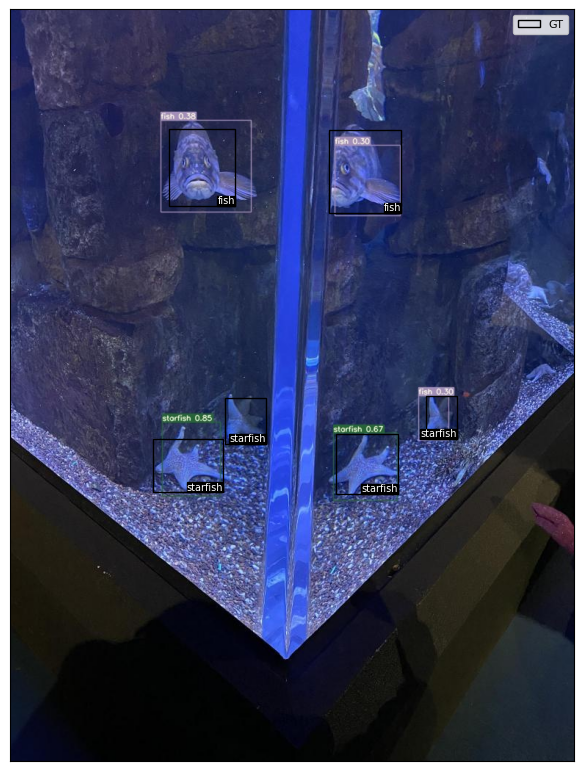}}
     
     \subfloat[(BCCD) KFold]{\includegraphics[width=0.48\textwidth]{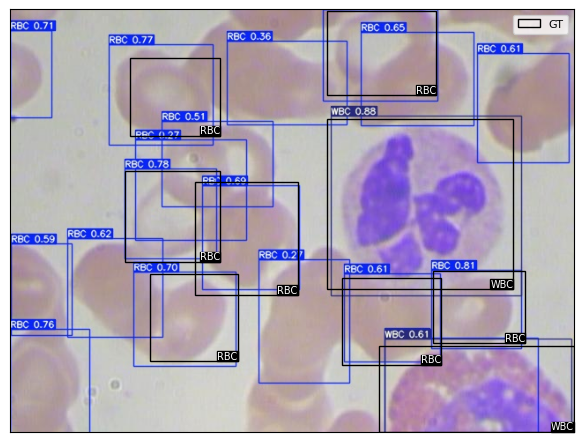}}
     \hfill
     \subfloat[(BCCD) Our method]{\includegraphics[width=0.48\textwidth]{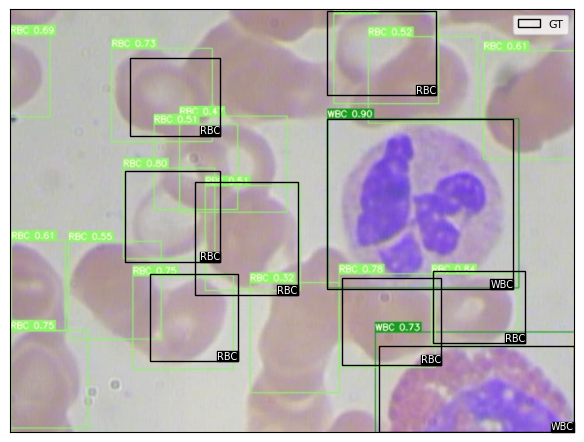}}
     \caption{Comparing KFold to our proposed method on the private dataset.}
\end{figure*}

In Figure 3, the inference results for the public dataset are presented. When comparing KFold and our method on the COCO dataset, the classification error for our method is less than that using KFold. While the tennis racket located in the center of the image is not accurately classified with KFold, it is correctly classified using our method. This demonstrates that using our method to split the dataset results in a smaller classification error as the class ratio is maintained throughout the dataset partitioning. A comparison conducted on the Pascal VOC 2012 dataset shows that the bounding box appearing in the upper left of the image is drawn incorrectly when using KFold but is drawn closer to the correct answer when using our method. This is also due to the reduction in localization error\cite{local_error} as the aspect ratio of the bounding box is maintained with the split of the dataset. For the PlantDoc dataset, using KFold resulted in two bounding boxes with different classes drawn in a similar location even though there is only one ground truth, leading to multiple detections. This occurs because the class ratios between the subsets are not maintained in the random splitting of the dataset with this method. Conversely, this phenomenon did not appear using our method.

In Figure 4, the inference results on the private dataset are shown. In the Website screenshot dataset, there is a True Negative\cite{tn} owing to the complexity of features within a class, but our method shows a higher IoU than does KFold. In this case, the localization error is reduced by maintaining the aspect ratio of the bounding box similar as before after splitting the dataset. In the Aquarium dataset, the True Negative phenomenon of no detection occurred when using KFold; however, this phenomenon was reduced when using our method. In the BCCD dataset, by optimizing the anchor box considering the class ratio, width, height, and ratio of the object, both KFold and our method achieved results closer to the ground truth of the bounding box and realized a higher IoU.

\section{Discussion $\&$ Conclusion}
This study focuses on the importance of applying stratification in the preprocessing stage of object detection tasks. This is the first study, to our knowledge, in the field of object detection that proposes a stratification method in dataset splitting, demonstrating the effect of alleviating class imbalance in the generation of training and validation sets. Additionally, stratification has been demonstrated to enhance the performance of an object detection model for a 2D image object detection dataset. However, stratification did not yield the best results under all experimental conditions. For high-dimensional datasets with a large number of classes, the application of stratification was ineffective, indicating that it may not be suitable for all types of datasets. Nevertheless, research involving classification tasks has shown that class balancing is essential and can significantly contribute to performance improvement for unbalanced datasets. This aspect is applicable to the object detection task in this study, emphasizing the need for an integrated approach to apply stratification in classification and object detection tasks. The results of this study confirm that the application of stratification must be carefully performed based on the characteristics of the dataset and distribution of the detection targets. As the features present in image data can influence the effect of stratification, further research in needed to improve and optimize the stratification method for use on images. Based on the results of this study, greater attention toward stratification utilization and its role in enhancing performance of object detection models is expected.

\authorcontributions{Conceptualization, H.L., S.A.; methodology, H.L.; software, H.L.; validation, H.L., S.A.; formal analysis, H.L.; data curation, H.-L.; writing---original draft preparation, H.L.; writing---review and editing, H.L., S.A.; visualization, H.L.; funding acquisition, S.A. All authors have read and agreed to the published version of the manuscript.}

\institutionalreview{Not applicable.}

\dataavailability{Datasets used in this study are publicly available on the web(\url{https://public.roboflow.com/}).}

\acknowledgments{This work was supported by the National Research Foundation of Korea(NRF) grant funded by the Korea government(MSIT)(No. NRF-2022R1A4A1023248)}

\conflictsofinterest{The authors declare no conflict of interest.}


\begin{adjustwidth}{-\extralength}{0cm}

\reftitle{References}

\end{adjustwidth}
\end{document}